# Q-VALUED NEURAL NETWORK AS A SYSTEM OF FAST INDENTIFICATION AND PATTERN RECOGNITION

D.I.Alieva, B.V.Kryzhanovsky, V.M.Kryzhanovsky[1], A.B.Fonarev[2]

[1]Institute of Optical Neural Technologies Russian Academy of Sciences
119333, Moscow, Vavilov street 44/2, (095)-135-7802, e-mail: *iont@iont.ru*

[2]Department of Engineering Science and Physics, The College of Staten Island
CUNY, SI, New York 10314

**Abstract**

An effective neural network algorithm of the perceptron type is proposed. The algorithm allows us to identify strongly distorted input vector reliably. It is shown that its reliability and processing speed are orders of magnitude higher than that of full connected neural networks. The processing speed of our algorithm exceeds the one of the stack fast-access retrieval algorithm that is modified for working when there are noises in the input channel.

## 1. Introduction

The analysis of parametric neural networks [3-8] has shown that at the present parametric vector models of the associative memory are the best both with regard to the storage capacity and noise immunity. In the same time such high parameters of the aforementioned models were not used up to now. The situation changed after the publication of the paper [5], where the algorithm of mapping of binary patterns into $q$-valued ones was proposed. It was also shown in [5] that such mapping allows one to use vector neural networks for storing and processing of signals of any type and any dimension. Moreover, the mapping brings to nothing the main difficulty of all the associative memory systems, which is the negative influence of correlations between the patterns. Thus, in the cited work the authors showed the effective and simple approach to use vector models of neural networks and all the advantages related with them.

Here we use the approach of [3-5] to create a fast-access retrieval algorithm to operate with long data lists in the case of large distortions in the input channel. We shall call this algorithm as *identifying perceptron*. The reliability and processing speed of identifying perceptron are orders of magnitude higher than that of full connected neural networks, as well as the stack fast-access retrieval algorithm that is modified for working when there are noises in the input channel.

## 2. The perceptron scheme

At first, let us examine the scheme of fast neural network search for the case of a great number of colored randomized patterns (multiparameter vectors). This scheme can be simple generalized on the case of vectors of an arbitrary dimension, in particular on binary vectors, and on strongly correlated patterns.

So, let us have a set of $N$-dimensional $q$-valued patterns $\{X_\mu\}$:

$$X_\mu = (\mathbf{x}_{\mu 1}, \mathbf{x}_{\mu 2}, ..., \mathbf{x}_{\mu N})$$

where $\mathbf{x}_{\mu i}$ is the unit vector directed along one of the Cartesian axes of $q$-dimensional space ($\mu = 0, 1, ..., M-1$; $i = 1, 2, ..., N$), that is $\mathbf{x}_{\mu i} \in \{\mathbf{e}_k\}^q$, where $\{\mathbf{e}_k\}^q$ is the set of basis vectors of the space $R^q$. Each pattern $X_\mu$ is one to one associated with its key (identifier) that is $q$-valued $n$-dimensional vector $Y_\mu$:

$$Y_\mu = (\mathbf{y}_{\mu 1}, \mathbf{y}_{\mu 2}, ..., \mathbf{y}_{\mu n}), \quad \mathbf{y}_{\mu i} \in \{\mathbf{e}_k\}^q$$

The number of its components $n = \log_q M + 1$ is enough for coding of the key of the given pattern. As a key we understand either a directive $Y_\mu$, produced as a result of the input $X_\mu$, or any other information related to the pattern $X_\mu$. For definiteness, let us define that the key is the number $\mu$ of the pattern $X_\mu$. In this case $Y_\mu$ is the coding of the number $\mu$: the sequence of numbers of the basis vectors of $q$-dimensional vector space along which the unit vectors $\mathbf{y}_{\mu 1}, \mathbf{y}_{\mu 2}, ..., \mathbf{y}_{\mu n}$ are directed is just the number $\mu$ in the q-nary presentation. For simplicity we suppose that the patterns $\{X_\mu\}$ are the set of colored images with a number of pixels $N$ and a number of different colors $q$. In this case with each color we associate an integer number ranging between 1 and $q$, or, equivalently, a unit vector in q-dimensional space.

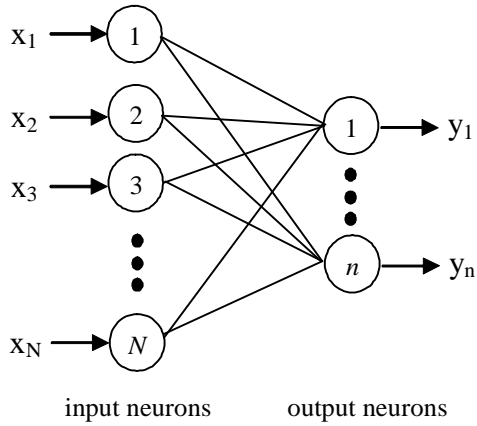

Fig.1. The scheme of the vector perceptron

Our goal is to create a neural network that is able to retrieve the key corresponding to the pattern, whose distorted image was presented to the network. The perceptron solving the problem is shown in Fig.1. It consists of two layers of vector neurons, where each neuron of the input layer is connected with all the neurons of the output layer. The algorithm works as follows. We construct the vector neural network based on standard patterns $\{X_\mu\}$ and $\{Y_\mu\}$ (Fig.1). The network consists of $N$ input and $n$ output neurons. To construct the network, we use the parametrical vector neuron approach [3-7]. Each neuron may be in one of $q$ different states, where $q \geq 2$. The $k$-th state of a neuron is associated with the basis vector $\mathbf{e}_k \in R^q$ of the $q$-dimensional space ($k = 0,1,...,q-1$). The same as in [3-5], the interconnection matrix elements are given by the generalized Hebb rule:

$$\mathbf{T}_{ij} = (1-\delta_{ij})\sum_{\mu=1}^{M} \mathbf{y}_{\mu i} \mathbf{x}_{\mu j}^{+} \qquad (1)$$

where $\mathbf{x}_{\mu j}^{+}$ is the vector-row ($1 \leq i \leq n$, $1 \leq j \leq N$), $\delta_{ij}$ is the Kronecker symbol. Note, in this approach the interconnections $\mathbf{T}_{ij}$ are not scalars, as they are in the standard Hopfield model, but $q \times q$ matrices.

### 3. Identification algorithm

Let an image $X = (\mathbf{x}_1, \mathbf{x}_2, ..., \mathbf{x}_N)$ be input to the network. Let it be a distorted copy of the $m$-th pattern $X_m \in \{X_\mu\}^M$. The local field created by all the neurons from the input layer, which is acting on the $i$-th output neuron, can be calculated as:

$$\mathbf{h}_i = \mathbf{h}_{i0} + \sum_{j=1}^{N} \mathbf{T}_{ij} \mathbf{x}_j \qquad (2)$$

where

$$\mathbf{h}_{0i} = -\frac{N}{q}\sum_{\mu=1}^{M} \mathbf{y}_{\mu i}$$

Under the action of the local field the $i$-th neuron becomes aligned along the basis vector, whose direction is the most close to that of the local field. The calculation algorithm is as follows:
1). The projections of the local field vector $\mathbf{h}_i$ onto all the basis vectors of the q-dimensional space are to be calculated;
2). The maximal projection is to be found. Let it be the projection onto a basis vector $\mathbf{e}_k$;
3). The value $\mathbf{y}_i = \mathbf{e}_k$ is set to the output neuron.

As it was shown in [5-9], under this dynamics all the components of the encoding output vector $Y_m$ are retrieved reliable.

Denote by $k_i$ the number of the basis vector along which the vector $\mathbf{y}_i$ is directed. Then the number of the standard $X_m$, whose distorted copy is the input vector $X$, is defined by the expression:

$$m = k_1 q^0 + k_2 q^1 + ... + k_n q^{n-1}$$

### 4. Characteristics of perceptron

To estimate the operational capability of the identifying perceptron, let us define its characteristics, namely the reliability of identification, the storage capacity and noise immunity, the number of output neurons and computational complexity of the algorithm.

The probability of the error in the pattern identification is defined by the expression

$$P = n\sqrt{\frac{M}{\pi N_e}} \exp\left(-\frac{N_e q^2}{4M}\right) \quad (3)$$

where $N_e = N(1-b)^2$, and $bN$ is the number of the distorted components of the input vector.

The maximal number of the standards $M_{max}$ which can be reliably identified by the identifying perceptron with the error probability less than a given value $P_0$, is defined from (3):

$$M_{max} = N_e \frac{q^2}{4|\ln P_0|} \quad (4)$$

As it is seen, the number of identified standards $M$ grows proportionally to the dimension $N$ and to the squared number of colors $q$. When the requirements to the recognition reliability become tougher, i.e. if the given value of the error $P_0$ decreases, the value of $M$ decreases logarithmically.

The critical distortion level at which the perceptron fails to identify patterns is defined by:

$$b_{max} \sim 1 - 2\sqrt{M}/q\sqrt{N} \quad (5)$$

Note, that if the number of patterns $M$ and their dimensionality $N$ are of the same order of magnitude, the critical distortion level can be rather high. For example, in Fig.2 we show the dependence of the retrieval reliability $\overline{P} = 1 - P$ on the parameter $b$ for $q = 8, 16, 32$ in the case $M/N = 2$. Above the threshold (5), the retrieval reliability decreases rapidly.

## 5. Discussion

We used the algorithm presented above to recognize and identify $N$-dimensional $q$-valued input patterns. However, rather easily it can be adapted for processing of binary input patterns by means of simple preprocessing, which is the mapping of the binary vector onto q-valued pattern (see [5]). After that we use our algorithm. Note, the mapping procedure suppresses the correlations between the input patterns [5], [9].

Comparing the expression (3) with the results of the works [1], [2], [10], [11], we see that for the perceptron algorithm the probability of the error recognition is $N$ times less than in the case of the full connected neural networks. Actually this means that our vector perceptron is able to identify reliably the input pattern, even when the full connected neural networks yield a priory incorrect output signals.

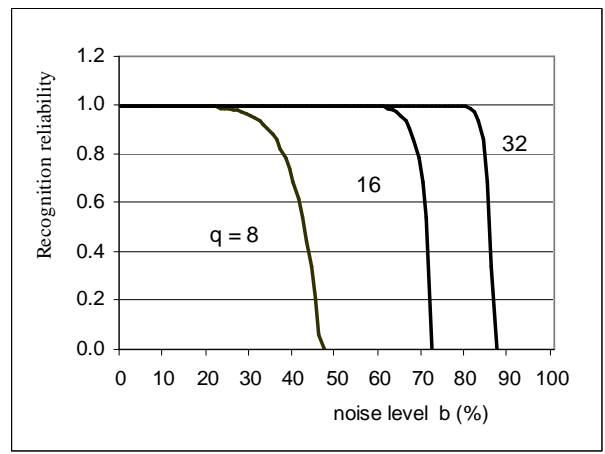

Fig.2 Recognition reliability as a function of noise level $b$.

The high parameters of the reliability of our algorithm allows one to use it for fast search in a long list, in particular when distortions are sufficiently large. For example, the aforementioned algorithm can identify any of $M \sim 10^5$ patterns with the reliability 0.99997 during the time of ~70ms. A pattern is an image of $N$=300x136 pixels, the number of different levels of brightness or different colors is equal to $q$=256, the number of the distorted pixels is up to 95%.

Let us estimate the calculation complexity of the algorithm. For identification of $N$-dimensional $q$-valued pattern $nNq$ operations are necessary. It is considerably less than the direct list sample, where $NM$ operations of comparison with respect to the proximity are necessary. Moreover, if rather low level of distortions about 15-20% is exceeded, our algorithm becomes preferable comparing with the well-known fast search algorithm based on the stack algorithm. The processing speed of the last decreases drastically, when the distortions of the recognized (identified) pattern increase, while the processing speed of our algorithm is independent of the distortion level [5]. The typical situation is illustrated in Fig. 3. Note, for large values of the parameter $q$ the perceptron algorithm works faster at any distortion level.

The number of output neurons $n$ is much less in comparison with the number of input neurons. It can be estimated as

$$n_{\max} = 2 + \ln N_e / \ln q$$

When $q$ is large enough, as a rule this quantity exceeds 3 slightly. For instance, in the aforementioned example four output neurons satisfy completely all the requirements of the algorithm.

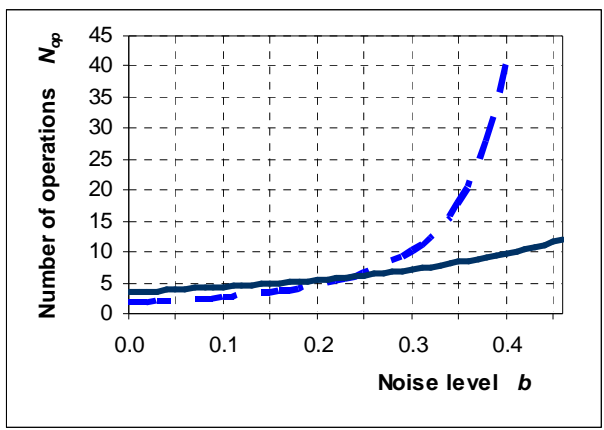

Fig.3 The dependence of the number of operations (x$10^3$) on the distortion level for perceptron algorithm (solid line) and for stack

The work was supported by grants of Russian Foundation of Basic Research (04-07-90038 и 03-01-00355) and project 1.8 of DITCS RAS.